\documentclass[conference]{IEEEtran}
\IEEEoverridecommandlockouts

\usepackage{cite}
\usepackage{amsmath,amssymb,amsfonts,bm}
\usepackage{algorithmic}
\usepackage{graphicx}
\usepackage{textcomp}
\usepackage{xcolor}

\usepackage{authblk}
\usepackage{subcaption}
\usepackage{gensymb}
\usepackage{url}
\usepackage{tikz,tikz-3dplot}
\usepackage{booktabs}

\def\BibTeX{{\rm B\kern-.05em{\sc i\kern-.025em b}\kern-.08em
    T\kern-.1667em\lower.7ex\hbox{E}\kern-.125emX}}

\newcommand{\name}[1]{AERIS}
    
\begin{document}

\title{AERIS: Argonne Earth Systems Model\\for Reliable and Skillful Predictions\thanks{Authors are listed alphabetically by their last names.}}

\author[1,+]{Väinö Hatanpää}
\author[1,+]{Eugene Ku}
\author[1,+]{Jason Stock}
\author[1]{Murali Emani}
\author[1]{Sam Foreman}
\author[1]{Chunyong Jung}
\author[1]{\\Sandeep Madireddy}
\author[4]{Tung Nguyen}
\author[1]{Varuni Sastry}
\author[2]{Ray A. O. Sinurat}
\author[1]{Sam Wheeler}
\author[1]{\\Huihuo Zheng}
\author[1,3]{Troy Arcomano}
\author[1,*]{Venkatram Vishwanath}
\author[1,*]{Rao Kotamarthi}

\affil[1]{Argonne National Laboratory, Lemont, Illinois, USA}
\affil[2]{University of Chicago, Chicago, Illinois, USA}
\affil[3]{Allen Institute for AI, Seattle, Washington, USA}
\affil[4]{University of California, Los Angeles, California, USA}
\affil[+]{\textit{Joint First Authors}$\quad^\mathrm{*}$\textit{\{venkat,vrkotamarthi\}@anl.gov}}

\maketitle

\begin{abstract}
Generative machine learning offers new opportunities to better understand complex Earth system dynamics. Recent diffusion-based methods address spectral biases and improve ensemble calibration in weather forecasting compared to deterministic methods, yet have so far proven difficult to scale stably at high resolutions. We introduce \name{}, a 1.3 to 80B parameter pixel-level Swin diffusion transformer to address this gap, and SWiPe, a generalizable technique that composes window parallelism with sequence and pipeline parallelism to shard window-based transformers without added communication cost or increased global batch size. On Aurora (10,080 nodes), \name{} sustains 10.21 ExaFLOPS (mixed precision) and a peak performance of 11.21 ExaFLOPS with 1${\times}$1 patch size on the 0.25$^\circ$ ERA5 dataset, achieving 95.5\% weak scaling efficiency, and 81.6\% strong scaling efficiency. \name{} outperforms the IFS ENS and remains stable on seasonal scales to 90 days, highlighting the potential of billion-parameter diffusion models for weather and climate prediction.
\end{abstract}
\vspace{1em}

\section{Justification} \label{sec:justification}
\name{} achieves a sustained mixed-precision performance of 10.21 ExaFLOPS and peak performance of 11.21 ExaFLOPS, scaling to 10,080 nodes (120,960 GPU-tiles) on the Aurora supercomputer---highest achieved to date in AI for Science. This accomplishment represents a significant breakthrough in generative modeling for science applications.


\section{Performance Attributes}
\noindent\begin{tabular}{{p{0.42\linewidth}p{0.5\linewidth}}}
\hline
Performance Attribute & Our Submission  \\ 
\hline
Category of achievement  & Scalability; time-to-solution \\
Type of method used      & Explicit; deep learning \\
Results reported with    & Whole application + I/O \\
Precision reported       & Mixed precision (\texttt{BF16})\\
System scale             & Measured on full system  \\
Measurement mechanism    & Timers; performance modeling  \\
\hline            
\end{tabular}
  
\begin{figure*}[!t]
  \centering
  \includegraphics[width=\linewidth]{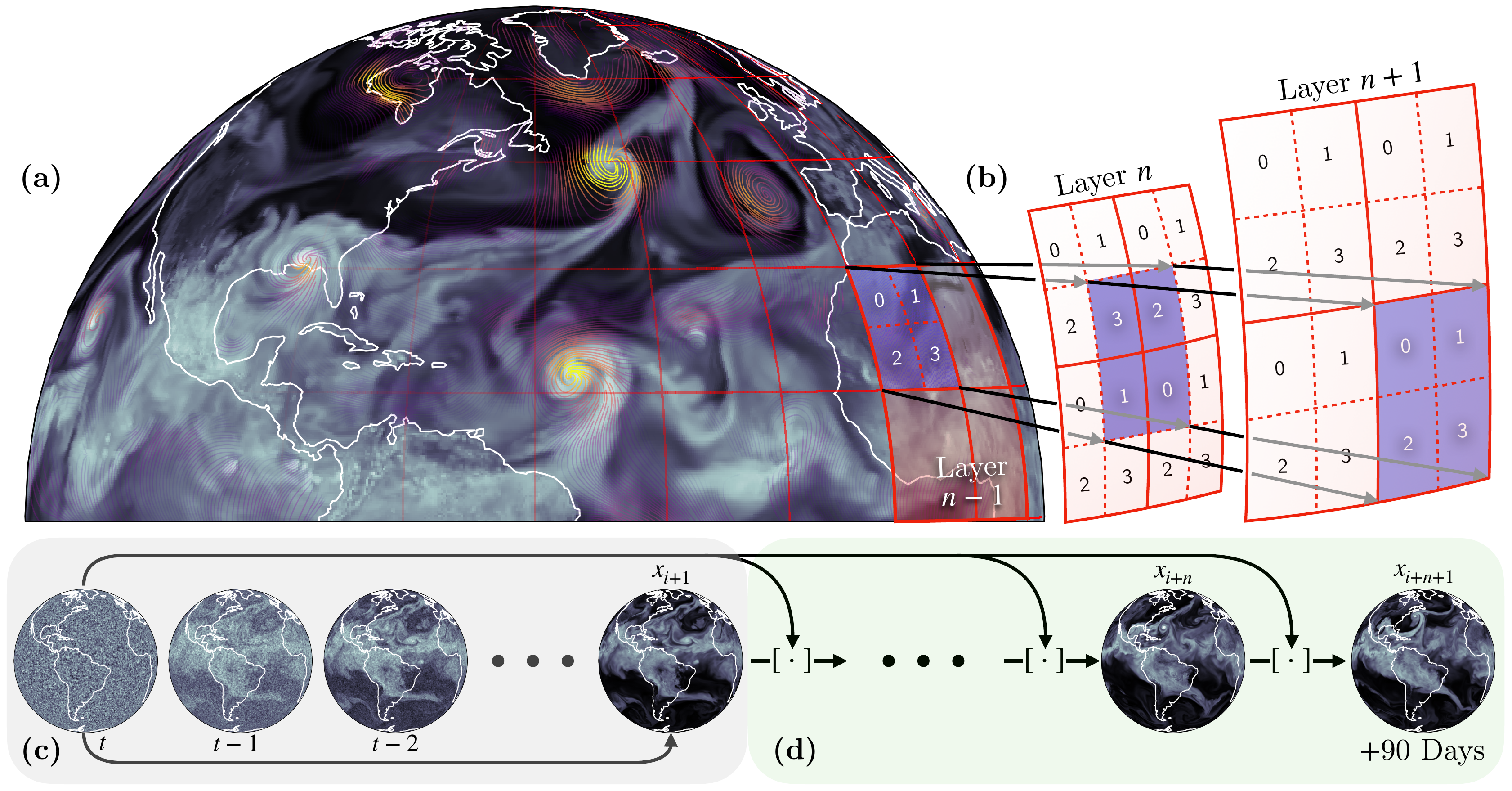}
  \caption{High-level overview of AERIS. (a) Forecast over the Atlantic basin showing Hurricane Teddy, Tropical Storm Sally, and Post-Tropical Cyclone Paulette seen with 10m wind tangents and intensity superimposed on specific humidity at 700 hPa (Q700; 2020-09-16T18z) with $60\times60$-pixel (at $15^\circ$) \textcolor{red}{windows outlined in red}; (b) sequence-window parallelism over network layers $n-1$ to $n+1$ optimally sliding partitions of an \textcolor{blue}{example window} between nodes; (c) iterative \textcolor{gray}{diffusion steps} to generate diverse global ensembles from any given initial condition; and (d) \textcolor{green!60!black}{autoregressive steps} to generate stable forecasts to 90 days.}
  \label{fig:cover}
\end{figure*}

\section{Overview of the Problem}

Weather and subseasonal-to-seasonal (S2S) forecasting is a fundamental problem for science and society. Accurate forecasts helps us prepare and recover from the effects of natural disasters and extreme weather events. Traditionally, domain scientists have relied on numerical weather prediction (NWP) techniques \cite{bauer2015quiet} to simulate and model complex atmospheric and climate dynamics, including both short- and long-term (on seasonal scale) weather forecasting. These models utilize systems of differential equations describing fluid flow and thermodynamics, which can be integrated over time to obtain future forecasts~\cite{lynch2008origins,bauer2015quiet}. When computed on large CPU-based HPC machines, these models typically produce global, 14-day forecasts four times a day \cite{wedi2015modelling}. 

Despite their widespread use, NWP models face several challenges. They suffer from parameterization uncertainties of important small-scale physical processes, including cloud physics and radiation, which can affect forecasting accuracy~\cite{stensrud2009parameterization}. They also incur high computation costs due to the complexity of numerical integration. Furthermore, NWP forecast accuracy does not inherently improve with increasing amounts of data; instead, their effectiveness heavily depends on domain experts continuously refining equations, parameterizations, and numerical algorithms~\cite{magnusson2013factors}. These challenges compound significantly with increased spatiotemporal resolutions and forecast dimensions, e.g., the number of ensembles and lead time of future forecasts.

To address these aforementioned challenges, there has been an increasing interest in data-driven approaches based on deep learning models for weather forecasting~\cite{gmd-11-3999-2018,scher2018toward,weyn2019can} (also see Section~\ref{subsec:deep_learning_methods}). The central idea involves parameterizing neural networks to predict future weather conditions using a vast amount of historical data, such as the ECMWF Reanalysis v5 (ERA5) dataset~\cite{hersbach2018era5,hersbach2020era5,rasp2020weatherbench,rasp2023weatherbench}. Once trained, these models generate forecasts within seconds, compared to the hours needed by typical NWP models. 

Early attempts relied on coarse reanalysis data ($\sim$400km) and produced forecasts that fell short of current NWP~\cite{rasp2021data,Arcomano_et_al_2020a,weyn2020improving,weyn2021sub,clare2021combining}. In later works using the native $0.25^\circ$ resolution of ERA5 ($\sim$30km), alongside advances in architectures (i.e., transformers and graph neural networks) and training strategies \cite{keisler2022forecasting,pathak2022fourcastnet,willard2024analyzingexploringtrainingrecipes,nguyen2023climax,bi2023accurate,lam2022graphcast,chen2023fengwu,chen2023fuxi}, performance became competitive with NWP. Similar challenges arose in natural language processing \cite{grattafiori2024llama,bai2023qwen,bi2024deepseek,kaplan2020scaling} and computer vision \cite{zhai2022scaling,dehghani2023scaling}, where breakthroughs inspired analogous efforts in weather and climate modeling \cite{nguyen2024scaling,wang2024orbitoakridgebase,han2024fengwu}. These studies show that higher horizontal and vertical resolution, together with architectural scaling, improves forecast skill, particularly with transformer-based models \cite{willard2024analyzingexploringtrainingrecipes,pathak2022fourcastnet,nguyen2024scaling}. However, these models remain expensive to train and are limited by existing parallelism techniques (e.g., sequence, pipeline, and sharded data parallelisms), limitations that compound with increasing model size and data resolution.

Among transformer-based architectures, the Swin Transformer \cite{liu2021swin,liu2022swin} is particularly well-suited for spatiotemporal modeling, combining the structured efficiency of window-based attention with the scaling advantages of standard transformers. Its design makes it effective for high-resolution data and compatible with existing parallelisms. At the same time, the independence of localized attention windows exposes an additional layer of parallelism that, despite its potential, has received little attention. In this work, we identify and exploit this windowed structure, significantly improving the scalability potential of window-based transformers.

To this end, we introduce \name{}, a pixel-level Swin Transformer for data-driven weather and S2S forecasting (Figure~\ref{fig:cover}). The architecture builds on advances from recent large-scale models and leverages our generalizable window-based parallelism strategy to scale efficiently with both model size and data resolution. Trained as a generative diffusion model on global $0.25^\circ$ ERA5 reanalysis data, our formulation produces 6- and 24-hourly forecast ensembles competitive on medium-range with impressive seasonal stability, using an architecture that can be finetuned or distilled to downstream tasks.

In summary, our contributions are: 
\begin{itemize}
    \item Stable large-scale training of 1.3B--80B parameter vision models, sustaining 10.21 ExaFLOPS in mixed-precision performance on 10,080 Aurora nodes (120,960 GPU-tiles) and 0.54 ExaFLOPS on 1,008 LUMI nodes.
    \item The first billion-parameter diffusion model for weather and climate, operating at the pixel level ($1\times 1$ patch size) and further guided by physical priors.
    \item SWiPe, a generalizable parallelism strategy that shards window-based transformers across high-resolution inputs, enabling efficient small-batch training and scalable performance on large supercomputers.
    \item Medium-range forecast skill surpassing IFS ENS and competitive with GenCast, while uniquely stable on seasonal scales to 90 days, validated through domain-specific diagnostics and case studies of hurricanes and heatwaves.
\end{itemize}
\section{State of the Art} \label{sec:current}

Section \ref{subsec:deep_learning_methods} begins with a review of related machine learning approaches for weather prediction and their relevance to this work, followed by a discussion in Section \ref{ssec:scaling-strategies} comparing existing parallelism techniques and scaling strategies.

\subsection{Deep Learning Methods \label{subsec:deep_learning_methods}}
There are many data-driven, machine learning approaches to model the global evolution of the Earth system. Broadly, these can be divided into deterministic methods \cite{lam2023graphcast,pathak2022fourcastnet,bi2022pangu,willard2024analyzingexploringtrainingrecipes,nguyen2024Stormer} and probabilistic or generative ones \cite{price2024gencast,couairon2024archesweather,stock2024diffobs,mardani2025residual}. The former is widely adopted in practice, owing to their relative training simplicity, with models such as GraphCast \cite{lam2023graphcast} and FourCastNet \cite{pathak2022fourcastnet} delivering competitive medium-range skill (5--10 days) at only a fraction of the cost of their numerical counterparts. However, their forecasts tend to underperform in longer term and ensemble settings, producing blurred, poorly calibrated distributions due to spectral biases \cite{ebert2025measuring,chattopadhyay2023challenges} and a lack of sensitivity to initial-condition perturbations \cite{butterfly_effect}. These shortcomings persist across network architecture, although transformer-based approaches show evidence of improved forecast skill proportional to the scale of larger parameter counts and smaller patch sizes \cite{nguyen2024Stormer,willard2024analyzingexploringtrainingrecipes}.

Advances in generative modeling, particularly with diffusion methods \cite{sohl2015deep,ho2020denoising,song2020score,lu2024simplifying,karras2022elucidating}, similarly show favorable scaling laws on computer vision tasks \cite{li2024scalability,peebles2023scalable,liang2024scaling}. Unlike deterministic methods, these models learn conditional probabilities, naturally quantify uncertainty, and are robust under incomplete distributional assumptions, making them well-suited to model the stochastic dynamics of the atmosphere. Recent diffusion-based weather models, such as GenCast \cite{price2024gencast} (used as a baseline in this work), better captures small-scale variability and produce ensembles approaching those of numerical systems, thereby addressing several limitations of deterministic models. 

Nonetheless, GenCast faces important challenges: its multi-step solver becomes unstable beyond two weeks at $0.25^\circ$ resolution, and its graph neural network backbone is less amenable to large-scale training compared to transformers. This raises a fundamental question of whether diffusion transformers, when scaled to billions of parameters, can achieve the stability and expressivity required for global, high-resolution weather and climate prediction. Our work addresses this gap, with a non-hierarchical, pixel-level Swin (window-based) transformer \cite{liu2021swin,liu2022swin} akin to \cite{willard2024analyzingexploringtrainingrecipes}, but parameterized by diffusion with architectural and parallelism innovations to improve stability, scalability, and long-range forecast skill.

\subsection{\label{ssec:scaling-strategies}Scaling Strategies}

State-of-the-art training methods for parallelizing large transformer models involve combining data parallelism, pipeline parallelism, tensor-parallelism, and sequence parallelism (i.e., 4D parallelism). Among them, \textbf{Data Parallelism (DP)} (e.g. DDP, FSDP, ZeRO \cite{zero,FSDP}) is the simplest approach and requires the least communication (except for ZeRo3), but is limited by its inability to parallelize beyond the batch size, and naively increasing the batch size can negatively affect training efficiency and convergence. On the other hand, \textbf{Pipeline Parallelism (PP)} (e.g. GPipe, 1F1B \cite{gpipe,pippy2022}) can parallelize a model with respect to the number of layers instead of batch size, but its efficiency is hindered by the bubble size (idle time) which ironically also requires a large batch size to minimize. \textbf{Tensor Parallelism (TP)} (e.g., \cite{shoeybi2019megatron,rasley2020deepspeed}) can parallelize a model through sharding the head-dimension of multi-head attention and is effective at sharding memory incurring from model states (parameters, optimizers, gradients), but is ineffective at sharding activation memory. In addition, TP incurs high communication overhead, limiting it only as an intra-node parallelism in practice. On the contrary to TP, \textbf{Sequence Parallelism (SP)} (e.g. Ulysses, Tensor-Sequence Parallelism, Ring-Attention \cite{deepspeedulysses, TP-SP, RA-SP}) are effective at sharding the activation memory but do not shard any model state memory.

Another approach is \textbf{Domain Parallelism} (e.g., PyTorch DTensor and NVIDIA PhysicsNeMo's ShardTensor) that shards inputs over devices across spatiotemporal dimensions and automatically issues the necessary halo exchanges. This primarily targets activation-heavy regimes and composes with DP for model-state sharding, with efficiency governed by operator locality and interconnect bandwidth/latency. However, performance degrades for non-local operations (e.g., global attention or normalization), where large communication overhead becomes unavoidable.
 
Previously, many of these parallelisms have been employed to train large transformer based weather forecasting models. Namely, ORBIT \cite{wang2024orbitoakridgebase} uses a combination of TP and FSDP to scale their vision transformer model up to 113B, achieving a peak throughput of $1.6$ EFLOPS. While such approaches alleviate memory pressure from model states, they remain ineffective at sharding activation memory, which is critical for high-resolution settings such as the global 0.25$\degree$ data used in our work. Consequently, scaling transformer-based weather models in both parameter count and sequence length (driven by patch size) exposes the limitations of traditional 4D parallelism due to large communication overhead, limited parallelism degree, and other various inefficiencies (e.g., high number of synchronization points in Ring-attention). To address these challenges, we introduce a new dimension of parallelism---\textbf{Window Parallelism (WP)}---that is agnostic to domain and broadly applicable to window-based transformers. Furthermore, we propose a communication ``merging'' optimization that amortizes synchronization cost, significantly reducing the overhead of our method; see Section \ref{subsec:swipe}.

\begin{figure}[!t]
    \centering
    \begin{subfigure}[b]{1\linewidth}
        \centering
        \hspace{3mm}
        \begin{tikzpicture}[every node/.style={font=\sffamily}, every path/.style={font=\sffamily}]
    \draw[red, thick] (0,0) -- (2,0) -- (2,2) -- (0,2) -- cycle;
    \draw[black, thin, dash pattern=on 1pt off 1pt] (0,0.25) -- (2,0.25);
    \draw[red, thick] (0,0.5) -- (2,0.5);
    \draw[black, thin, dash pattern=on 1pt off 1pt] (0,0.75) -- (2,0.75);
    \draw[red, thick] (0,1) -- (2,1);
    \draw[black, thin, dash pattern=on 1pt off 1pt] (0,1.25) -- (2,1.25);
    \draw[red, thick] (0,1.5) -- (2,1.5);
    \draw[black, thin, dash pattern=on 1pt off 1pt] (0,1.75) -- (2,1.75);

    \draw[black, thin, dash pattern=on 1pt off 1pt] (0.25,0) -- (0.25,2);
    \draw[red, thick] (0.5,0) -- (0.5,2);
    \draw[black, thin, dash pattern=on 1pt off 1pt] (0.75,0) -- (0.75,2);
    \draw[red, thick] (1,0) -- (1,2);
    \draw[black, thin, dash pattern=on 1pt off 1pt] (1.25,0) -- (1.25,2);
    \draw[red, thick] (1.5,0) -- (1.5,2);
    \draw[black, thin, dash pattern=on 1pt off 1pt] (1.75,0) -- (1.75,2);

    \draw[black, thick] (2.75,0) -- (4.75,0) -- (4.75,2) -- (2.75,2) -- cycle;
    \draw[black, thin, dash pattern=on 1pt off 1pt] (2.5,0.25) -- (4.75,0.25);
    \draw[black, thin, dash pattern=on 1pt off 1pt] (2.5,0.75) -- (4.75,0.75);
    \draw[black, thin, dash pattern=on 1pt off 1pt] (2.5,1.25) -- (4.75,1.25);
    \draw[black, thin, dash pattern=on 1pt off 1pt] (2.5,1.75) -- (4.75,1.75);
    \draw[black, thin, dash pattern=on 1pt off 1pt] (2.5,0.5) -- (4.75,0.5);
    \draw[black, thin, dash pattern=on 1pt off 1pt] (2.5,1) -- (4.75,1);
    \draw[black, thin, dash pattern=on 1pt off 1pt] (2.5,1.5) -- (4.75,1.5);
    \draw[black, thin, dash pattern=on 1pt off 1pt] (3,0) -- (3,2.25);
    \draw[black, thin, dash pattern=on 1pt off 1pt] (3.5,0) -- (3.5,2.25);
    \draw[black, thin, dash pattern=on 1pt off 1pt] (4,0) -- (4,2.25);
    \draw[black, thin, dash pattern=on 1pt off 1pt] (4.5,0) -- (4.5,2.25);
    \draw[black, thin, dash pattern=on 1pt off 1pt] (3.25,0) -- (3.25,2.25);
    \draw[black, thin, dash pattern=on 1pt off 1pt] (3.75,0) -- (3.75,2.25);
    \draw[black, thin, dash pattern=on 1pt off 1pt] (4.25,0) -- (4.25,2.25);

    \draw[black, thin, dash pattern=on 1pt off 1pt] (2.5,2) -- (4.5,2);
    \draw[blue, thick] (2.5,0.25) -- (4.5,0.25);
    \draw[blue, thick] (2.5,0.75) -- (4.5,0.75);
    \draw[blue, thick] (2.5,1.25) -- (4.5,1.25);
    \draw[blue, thick] (2.5,1.75) -- (4.5,1.75);

    \draw[black, thin, dash pattern=on 1pt off 1pt] (2.75,0.25) -- (2.75,2.25);
    \draw[blue, thick] (4.5,0.25) -- (4.5,2.25);
    \draw[blue, thick] (3,0.25) -- (3,2.25);
    \draw[blue, thick] (3.5,0.25) -- (3.5,2.25);
    \draw[blue, thick] (4,0.25) -- (4,2.25);

    \draw[black, thick] (-3,0) -- (-1,0) -- (-1,2) -- (-3,2) -- cycle;
    \draw[black, semithick, dashed] (0,1.5) -- (-0.33,1);
    \draw[black, semithick, dashed] (-0.8,0.3) -- (-1,-0.02);
    \draw[black, semithick, dashed] (0,2) -- (-1,2);

    \draw[black, thick] (-2,0) -- (-2,2);
    \draw[black, thick] (-3,1) -- (-1,1);
    \draw[fill=black, opacity=0.25] (-3,1/3) -- (-1,1/3) -- (-1,2/3) -- (-3,2/3) -- cycle;
    \draw[fill=black, opacity=0.25] (-3,4/3) -- (-1,4/3) -- (-1,5/3) -- (-3,5/3) -- cycle;
    
    \draw[black, semithick, -latex] (5.1,1.9) -- (4.7,2.3);
    \draw[black, semithick, -{>[scale=0.75]}] (0.75,0.25) to[bend left=15] (-0.15,0.1);
    \draw[black, semithick, -{>[scale=0.75]}] (0.25,0.25) to[bend right=15] (-0.15,0.35);
    \draw[black, semithick, -{>[scale=0.75]}] (0.75,0.75) to[bend left=15] (-0.15,0.6);
    \draw[black, semithick, -{>[scale=0.75]}] (0.25,0.75) to[bend right=15] (-0.15,0.85);

    \node at (-0.5,0.85) {\tiny $(0,0)$};
    \node at (-0.5,0.6) {\tiny $(1,0)$};
    \node at (-0.5,0.35) {\tiny $(0,1)$};
    \node at (-0.5,0.1) {\tiny $(1,1)$};

    \node at (-2.5,1.83) {\tiny $rank\;0$};
    \node at (-2.5,1.5) {\tiny $1$};
    \node at (-2.5,1.17) {\tiny $2$};
    \node at (-2.5,0.84) {\tiny $6$};
    \node at (-2.5,0.51) {\tiny $7$};
    \node at (-2.5,0.18) {\tiny $8$};
    \node at (-1.5,1.83) {\tiny $3$};
    \node at (-1.5,1.5) {\tiny $4$};
    \node at (-1.5,1.17) {\tiny $5$};
    \node at (-1.5,0.84) {\tiny $9$};
    \node at (-1.5,0.51) {\tiny $10$};
    \node at (-1.5,0.18) {\tiny $11$};

    \node at (-2,2.5) {\scriptsize Attn. Window};
    \node at (1,2.5) {\scriptsize Image Crop};
    \node at (3.6,2.5) {\scriptsize Window Shift};
    
\end{tikzpicture}
        \caption{Window distribution. (left) window divided across SP ranks; (middle) attention grid with windows distributed in window parallel groups; and (right) shifted attention (\textcolor{blue}{in blue}) for alternating layers.}
        \label{fig:wpsp}
    \end{subfigure}
    \begin{subfigure}[b]{1\linewidth}
      \centering
      \begin{tikzpicture}[every node/.style={font=\sffamily}, every path/.style={font=\sffamily}]
    \draw[black, dashed, thick, rounded corners=5pt] (-3.5,0) -- (4,0) -- (4,2.5) -- (-3.5,2.5) -- cycle;
    \draw[black, dashed, thick, rounded corners=5pt] (-3.5,-1.5) -- (4,-1.5) -- (4, -0.65) -- (-3.5,-0.65) -- cycle;

    \node[rotate=90] at (-3.8,1.25) {\scriptsize \textbf{DP 0}};

    \node[rotate=90] at (-3.8,-1.075) {\scriptsize \textbf{DP 1}};
    
    \node at (0.125,-0.35) {\scriptsize Gradient Sync};
    \draw[black, semithick, latex-latex] (-1.25,-0.65) -- (-1.25,0);
    \draw[black, semithick, latex-latex] (1.5,-0.65) -- (1.5,0);
    \node at (0.125,-1.1) {\huge $\cdots$};

    \draw[red, thick] (-3,0.5) -- (-2.5,0.5) -- (-2.5,1.0) -- (-3,1.0) -- cycle;
    \draw[red, thick] (-3.1,0.6) -- (-2.6,0.6) -- (-2.6,1.1) -- (-3.1,1.1) -- cycle;
    \draw[red, thin, dash pattern=on 1pt off 1pt] (-3,0.75) -- (-2.5,0.75);
    \draw[red, thin, dash pattern=on 1pt off 1pt] (-2.75,0.5) -- (-2.75,1.0);
    \draw[black, semithick, -latex] (-2.5,0.75) -- (-2,0.75);
    \node[scale=0.6] at (-2.76, 0.76) {\small W$_{\text{N}}$};
    
    \draw[red, thick] (-3,1.5) -- (-2.5,1.5) -- (-2.5,2.0) -- (-3,2.0) -- cycle;
    \draw[red, thick] (-3.1,1.6) -- (-2.6,1.6) -- (-2.6,2.1) -- (-3.1,2.1) -- cycle;
    \draw[red, thin, dash pattern=on 1pt off 1pt] (-3,1.75) -- (-2.5,1.75);
    \draw[red, thin, dash pattern=on 1pt off 1pt] (-2.75,1.5) -- (-2.75,2.0);
    \draw[black, semithick, -latex] (-2.5,1.75) -- (-2,1.75);
    \node[scale=0.6] at (-2.76, 1.76) {\small W$_0$};

    \draw[red, thick] (3.0,0.5) -- (3.5,0.5) -- (3.5,1.0) -- (3.0,1.0) -- cycle;
    \draw[red, thick] (2.9,0.6) -- (3.4,0.6) -- (3.4,1.1) -- (2.9,1.1) -- cycle;
    \draw[red, thin, dash pattern=on 1pt off 1pt] (3.0,0.75) -- (3.5,0.75);
    \draw[red, thin, dash pattern=on 1pt off 1pt] (3.25,0.5) -- (3.25,1.0);
    \draw[black, semithick, -latex] (2.5,0.75) -- (2.9,0.75);
    \node[scale=0.6] at (3.24, 0.76) {\small W$_{\text{N}}$};

    \draw[red, thick] (3.0,1.5) -- (3.5,1.5) -- (3.5,2.0) -- (3.0,2.0) -- cycle;
    \draw[red, thick] (2.9,1.6) -- (3.4,1.6) -- (3.4,2.1) -- (2.9,2.1) -- cycle;
    \draw[red, thin, dash pattern=on 1pt off 1pt] (3.0,1.75) -- (3.5,1.75);
    \draw[red, thin, dash pattern=on 1pt off 1pt] (3.25,1.5) -- (3.25,2.0);
    \draw[black, semithick, -latex] (2.5,1.75) -- (2.9,1.75);
    \node[scale=0.6] at (3.24, 1.76) {\small W$_0$};

    \foreach \x/\y/\labelX in {
        -1.75/1.35/-1.25,  
        -1.75/0.15/-1.25,  
         1/0.15/1.5,       
         1/1.35/1.5        
    } {
        \draw[blue, thick] (\x,\y) rectangle ({\x+1}, {\y+1});
        
        \foreach \i/\name/\dy/\dx/\adx in {
            0/{GPU $\;0\;\;$}/0.75/0.125/-0.75,
            1/{GPU $\cdots$}/0.50/0.125/0.5,
            2/{GPU $\;$N$\;\,$}/0.25/0.125/0.25
        } {
            \pgfmathsetmacro{\commslineY}{\y+\dy}
            \draw[blue] (\x,\commslineY) -- ({\x+1},\commslineY);
            \node[scale=0.6] at (\labelX, {\commslineY+0.125}) {\small \name};
            \draw[gray, semithick, -latex] ({\x+1}, {\commslineY+\dx}) -- ({\x+1.4}, {\commslineY+\adx});
        }
    
        \filldraw[blue, opacity=0.25] (\x,\y) rectangle ({\x+1}, {\y+0.25});
        \node[scale=0.6] at (\labelX, {\y+0.125}) {\small SP};
    }
    \foreach \x/\y in {
         1/0.15,       
         1./1.35       
    } {
        \foreach \dy/\dx/\adx in {
            0.75/0.125/0.2,
            0.50/0.125/0.45,
            0.25/0.125/0.65
        } {
            \draw[gray, semithick, -latex] ({\x-0.4}, {\y+\adx}) -- ({\x}, {\y+\dy+\dx});
        }
    }

    \node at (0.125,1.75) {\scriptsize \textbf{WS+}};
    \node at (0.1,1.3) {\scriptsize \textbf{PP}};
    \node at (0.125,0.95) {\scriptsize \textbf{send /}};
    \node at (0.125,0.73) {\scriptsize \textbf{recv}};

    \node at (-1.3,2.7) {\scriptsize Stage / Layer $(n)$};
    \node at (1.5,2.7) {\scriptsize Stage / Layer $(n+1)$};

\end{tikzpicture}  
      \caption{Parallelism stages and synchronization}
      \label{fig:global_parallelisms.dp}
    \end{subfigure}
    
    \caption{High-level parallelism architecture of SWiPe}
    \label{fig:global_parallelisms}
\end{figure}
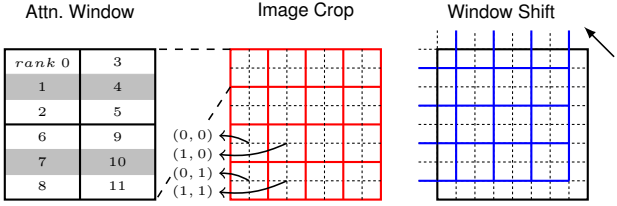
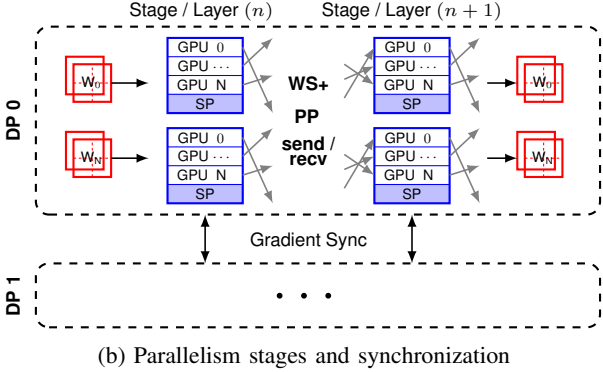

\section{Innovations Realized}
\label{sec:innovation}
We highlight the key innovations in this work, starting with our novel parallelism in Section \ref{subsec:swipe}, followed by the \name{} model architecture in Section \ref{subsec:model-architecture}.

\subsection{\label{subsec:swipe}SWiPe: Sequence-Window Parallelism}

\textbf{Overview } Our proposed Window Parallelism (WP) strategy exploits the inherent structure of Swin Transformers, which naturally partitions the computations across non-overlapping spatial windows. In Swin, tokens are grouped into distinct attention windows, enabling localized attention within each window while maintaining a larger receptive field across layers by shifting these windows. This design preserves global context without requiring global attention in each layer. Unlike other localized attention mechanisms, such as sliding window attention or graph-based transformers, Swin's grid-based attention provides a clean and effective way to partition the input image across multiple ranks. Each rank handles a disjoint set of attention windows, enabling parallel computation without requiring halo exchange or duplication of overlapping data. This property makes Swin particularly amenable to WP, allowing \name{} to scale model parallelism efficiently by assigning separate windows to different GPUs while minimizing the communication overhead.

We adopt a hierarchical parallelism scheme for WP. The attention windows of a single image are first distributed across a 2D compute node grid of $A \times B$. The assigned window on each node is further partitioned across all the GPUs within that node using Ulysses's Sequence Parallelism (SP). The parallelism strategy is further enhanced by combining WP with other existing parallelism schemes such as Pipeline Parallelism (PP). Our resulting hybrid hierarchical parallelism scheme, termed \textbf{Sequence-Window Parallelism (SWiPe)}, significantly extends the scaling limit of large scale window-based model training. This is important for high-resolution global weather and climate modeling, where synoptic-scale systems and embedded mesoscale extremes arise from fine-grid interactions. SWiPe preserves pixel-level, shifted-window attention across the globe without the prohibitive cost of full global attention, while aligning compute with locality and minimizing cross-rank communication, and thereby enabling efficient, domain-guided scaling that sustains forecast fidelity.

\textbf{Details } Figure~\ref{fig:wpsp} illustrates an input image that is partitioned across the 12 GPU-tiles within an Aurora compute node (system details in Table~\ref{tab:system-table}). The image is first divided into 4 spatial quadrants (a $2 \times 2$ grid) to accommodate the window shift. Accordingly, the tiles are logically arranged in a $2 \times 2 \times 3$ topology. The quadrants are one-to-one mapped to the 4 groups of tiles (each group contains 3 tiles). Each quadrant is sharded across 3 GPU tiles in that group. This method of data distribution primarily aims to balance computational load and minimize data movement during window shifting. In the context of SP compute and communication, however, the input tokens are flattened into a 1D sequence allowing efficiently gathering and redistribution across GPUs using \texttt{alltoall} communication both before and after the attention layer. This flattened representation simplifies the data exchange.

The windows are distributed across all the ranks in the WP group in a round-robin fashion in both X and Y directions as shown in the middle of Figure~\ref{fig:wpsp}. This distribution scheme allows batch processing of windows during window shifting leading to improved throughput; it also substantially simplifies the send/receive data movement pattern from one stage to another in PP while shifting windows. Each rank will send $1/SP$ of the window to the receiving rank in the next stage. No data redistribution is needed among the ranks in the next stage after receiving the window. If no WP is used, each rank will send an entire window and redistribution of data is needed among the ranks in the next stage to achieve the window shifting. A high-level illustration of how the parallelism strategies are integrated is shown in Figure~\ref{fig:global_parallelisms.dp}. To reduce communication overhead, SP groups are confined within individual nodes, enabling the frequent and bandwidth-intensive \texttt{alltoall} collectives to fully leverage the high-speed intra-node interconnect.


\textbf{Communication overhead } The primary communication overhead arises from intra-node \texttt{alltoall} communication related to SP or WP, internode \texttt{send/recv} communication due to PP, and the \texttt{allreduce} operation from data parallelism. The message size for both \texttt{alltoall} and \texttt{send/recv} communications is defined as $M = b\times s\times h / SP/WP$, where $b$ is the batch size, $s$ is sequence length, and $h$ is the hidden dimension. Introducing WP reduces message size associated with \texttt{alltoall} and \texttt{send/recv} communications, while the overhead from gradient \texttt{allreduce} remains unchanged. Overall, enabling WP decreases the communication load per device. The \texttt{send/recv} communications can also overlap with computation, just like in regular PP, potentially hiding most inter-node communication. 

Compared to input sharding with domain parallelism, which requires multiple re-sharding points for the Swin transformer, SWiPe avoids introducing additional communication or synchronization points. It is also fully compatible with efficient attention kernels, as tokens that attended to one another are implicitly gathered. In contrast, DTensor-based sharding demands explicit gathering of tokens for localized attention.

\textbf{Activation memory} When WP is enabled on top of both SP and PP, the activation memory is reduced by a factor of $WP$, thus reducing the need for activation checkpointing, which usually introduces additional recomputation of about $1/3$ of the total computation amount \cite{reducing-act}.

\textbf{Data loading } In WP, both the input and output are spatially partitioned so that each node loads only the data it processes. Only the first and last stages of the pipeline perform data loading and writing, respectively. With a WP group size of 16, each participating node handles just $1/16^{\text{th}}$ of the full image. This is particularly beneficial for high-resolution weather and climate datasets, where the size can be prohibitively large. Leveraging data formats that support efficient slicing (e.g., HDF5) allows each node to load only its required spatial windows, drastically reducing I/O per node and distributing the load evenly. Relative to configurations without WP, where full images are read redundantly or re-partitioned after loading, this strategy minimizes I/O overhead. In practice, the reduced and parallelized I/O is fully overlapped with the warm-up phase of the pipeline schedule, adding no training latency.

\textbf{Mixed precision } All compute-intensive operations, including matrix-matrix multiplications and the Flash Attention kernel, were performed using \texttt{BF16}, which is standard in large-scale deep learning workloads. To ensure numerical stability, components such as embeddings, primary gradients, model parameters, and gradient reductions were maintained in \texttt{FP32} (single-precision floating point). These operations account for only a small fraction of the total computational workload and were not measured separately. Empirically, we observed no significant throughput improvement when these components were cast to lower-precision formats, and thus retained them in \texttt{FP32} for consistency and stability.

We identify the following advantages of SWiPe, each of which are critical for large-scale climate model training:
\begin{itemize}
\item It \textbf{enhances} the degree of parallelism, enabling efficient training of models at larger scales.
\item It \textbf{reduces} reliance on data parallelism for scaling, potentially improving model convergence behavior.
\item It \textbf{decreases} communication overhead, potentially lead to better scaling efficiency. 
\item It \textbf{lowers} activation memory usage, potentially eliminating the need for activation checkpointing and reducing the overall memory footprint.
\end{itemize}

\begin{figure}[!hb]
  \centering
  \includegraphics[width=\linewidth]{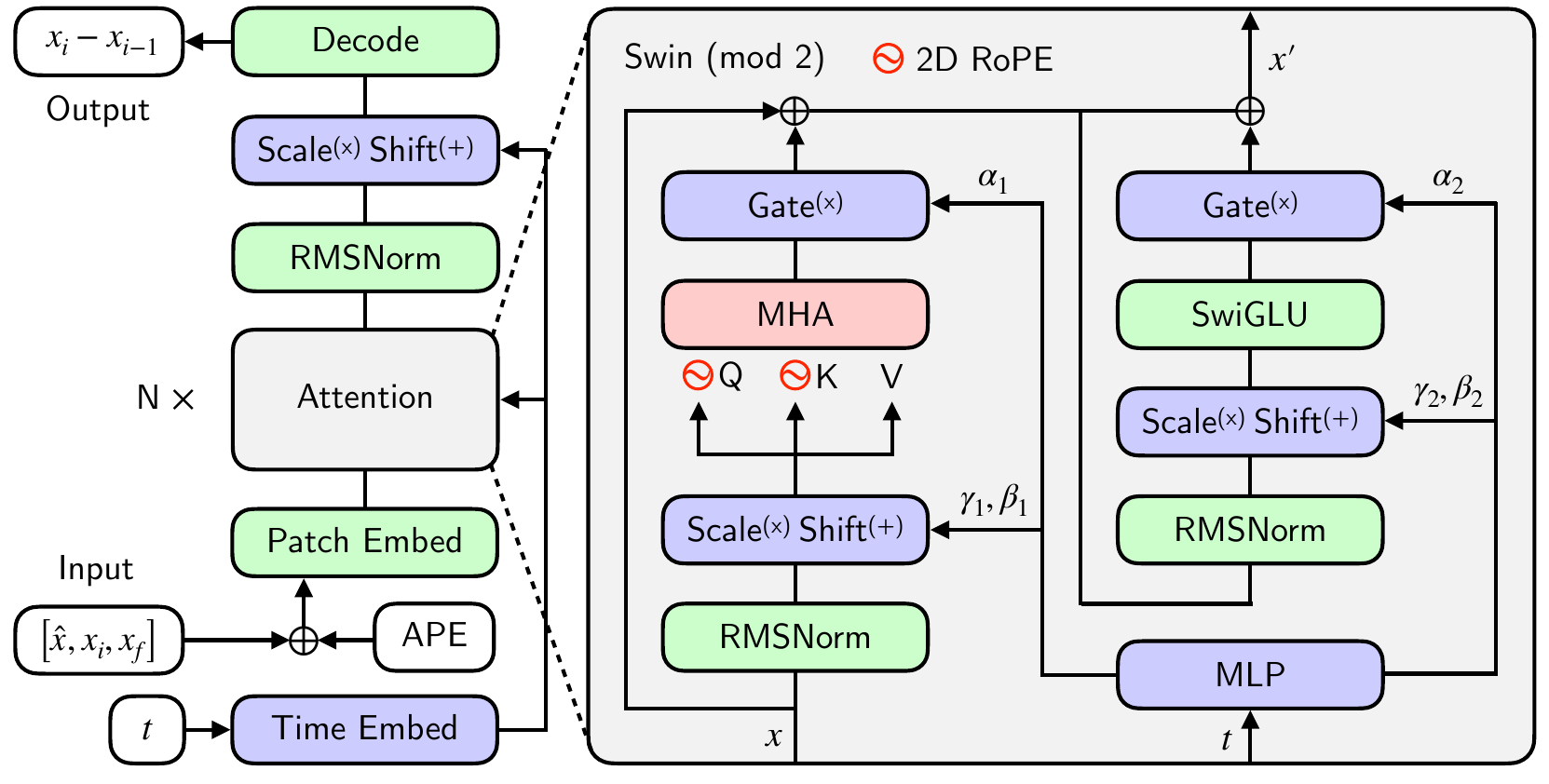}
  \caption{Model architecture.}
  \label{fig:model_arch}
\end{figure}

\subsection{\label{subsec:model-architecture}\name{} Model Architecture}
The Swin Transformer \cite{liu2021swintransformerhierarchicalvision,liu2022swin} is a prominent adaptation of the Vision Transformer \cite{dosovitskiy2021imageworth16x16words} that has demonstrated strong performance across a variety of computer vision applications \cite{willard2024analyzingexploringtrainingrecipes} despite lacking the global attention across all tokens. Instead, Swin shifts the receptive field of local attention windows every layer, effectively mimicking the global attention while obtaining the inductive bias of spatial-locality. Moreover, the lack of global attention affords greater flexibility of sequence length for attention, avoiding the costly quadratic compute complexity and enabling pixel-level patch-size.

The original Swin architecture uses hierarchical attention through shifted and downsampled windows for use in classification tasks. However, in this work, we map full resolution images in pixel-space, under a diffusion objective (see Section~\ref{ssec:application} for details), while maintaining a non-hierarchical structure, known to be beneficial for spatiotemporal tasks \cite{gao2023}. Specifically, \name{} is designed to be used autoregressively, in both diffusion and data steps (Figure~\ref{fig:cover}), taking as input a sample at time $i-1$ through $T$ diffusion steps to estimate the residual of a sample at time $i$.

In addition to its non-hierarchical structure, \name{} introduces several modifications that deviate from the original Swin Transformer. While SwinV2 \cite{liu2022swin} aimed to enhance training stability and dynamics for larger-scale models, our implementation builds upon this goal by incorporating modern techniques such as pre RMSNorm \cite{zhang2019root} and SwiGLU \cite{shazeer2020glu}, inspired by state-of-the-art large language models---particularly the Llama 3 series \cite{grattafiori2024llama}. 

Figure \ref{fig:model_arch} illustrates the conceptual flow of information and architecture, which is computationally optimized by our parallelism framework (Section~\ref{subsec:swipe}). We begin with adding a 2D sinusoidal positional encoding \cite{wang2021translating} to each channel of our input to serve as a proxy of locality given the spatial domain of our data. This input is embedded into an abstract representation through a learned linear layer before it passes through $N$ Swin Layers. Each layer is composed of multiple transformer layers consisting of pre RMSNorm (in place of LayerNorm) and uses SwiGLU (in place of a singular linear layer) in the fully-connected layers. Prior to the multi-headed attention block, we ``partition'' the embedded images into $30\times30$ (for 6h model) or $60\times60$ (for 24h model) windows, that are ``shifted'' every other layer. These are then classically projected to queries, keys, and values before dot product attention, where the queries and keys are projected via axial frequency 2D rotary positional embeddings \cite{heo2024rotary} (in place of relative positional biases). For the Attention we utilize the Ulysses sequence parallelism which does an \texttt{all-to-all} collective before and after the attention kernel. 
Following the last Swin Layer is a simple normalization and decoding block to project our embeddings back to pixel-space. 

The time embedding for the diffusion timestep is projected through a shared linear layer, and then further broadcasted to all the layers, which contain another layer-specific linear layer. The output of this linear layer is used as the values $\alpha$,$\beta$,$\gamma$ for the adaptive layer norm \cite{perez2018film,peebles2023scalable}.
\section{How Performance was Measured}
We describe the Aurora and LUMI systems used in the evaluation study, the application benchmark, the software stack and environment, and how we measure the performance.

\subsection{System Details}
Aurora is an exascale-class supercomputer hosted at the Argonne Leadership Computing Facility (ALCF). It is among the most powerful systems built~\cite{TOP500}, representing a major step forward in computational capabilities. 
We evaluate performance of training \name{} on Aurora. Table~\ref{tab:system-table} describes the key architectural characteristics of this HPE Cray EX system with 10,624 nodes interconnected with HPE Slingshot 11 using a Dragonfly topology. Each node consists of two Intel Sapphire Rapids processors with a total of 1024~GB of system memory. Each node has six Intel Data Center Max 1550 GPUs, each with 128~GB memory. A GPU has two compute tiles. Each node has eight Slingshot-11 endpoints at 25~GB/s each for the interconnect network. The GPUs are configuration in a standard mode with 896 execution units. Each GPU is capable of achieving a peak of 45 TFLOPS in \texttt{FP32}, 229 TFLOPS in TF32, and 458 TFLOPS in \texttt{FP16} and \texttt{BF16}.

LUMI is a petascale supercomputer hosted by LUMI consortium, located at CSC datacenter in Kajaani, Finland. LUMI is a HPE Cray EX system, with the GPU-partition consisting of 2978 nodes. The system is interconnected with HPE Slingshot 11 using a Dragonfly topology. Each node has four Slingshot-11 endpoints at 25 GB/s each for the interconnect network. Each node has four AMD MI250X GPUs, each containing two graphic compute dies (GCD). Each MI250X GPU has 128GB of memory. Each AMD MI250X GPU is capable of achieving a peak of 95.7 TFLOPS in \texttt{FP32} and 383 TFLOPS in \texttt{FP16} and \texttt{BF16}.

 \begin{table}[tb!]
\scriptsize
\caption{System configuration for performance evaluations.}
\centering
\begin{tabular}{r|c|c}
\toprule
                            & \textbf{Aurora} & \textbf{LUMI}    \\ 
\midrule
GPU                         & Intel Max 1550 & AMD MI250X  \\ 
GPUs (tiles) / node         & $6$($12$) & $4$($8$)          \\ 
GPU Memory (GB)             & $128$   & $128$             \\ 
GPU Memory Technology       & HBM2e & HBM2e           \\ 
GPU Memory BW (TB/s)        & $2.0$   & $3.2$    \\  
\midrule
Scale-Out Interconnect      & HPE Slingshot 11  & HPE Slingshot 11 \\ 
NICs / node                 & $8$ & $4$  \\ 
Network BW / direction (GB/s)    & $200$  & $100$ \\ 
Scale-up Interconnect            & All-to-All $X^e$ Links & Infinity Fabric \\ 
Scale-up  BW / direction (GB/s)  & $28$  & $50$ \\ 
Collective Communication Library & oneCCL & RCCL \\ 
\midrule
\textbf{Total nodes (GPU-tiles) scaled}   & $10{,}080$ ($120{,}960$) & $1{,}008$ ($8{,}064$)   \\ 
\bottomrule
\end{tabular}
\label{tab:system-table}
\end{table}

\subsection{Application Benchmark}\label{ssec:application}
\textbf{Dataset} We model the global evolution of the atmosphere, capturing medium-range and seasonal scales, by learning the state evolution $p(x_{i+1} | x_i)$ given four decades of ERA5 reanalysis data from the European Center for Medium-Range Weather Forecasting (ECMWF) \cite{hersbach2020era5}, as provided by WeatherBench2 (WB2)~\cite{rasp2024weatherbench2benchmarkgeneration,WB2}. Data are on the native 0.25$^\circ$ (720$\times$1440 pixel-grid with poles removed) spatial resolution with 6-hourly samples. We predict five surface-level variables: 2-meter temperature (T2m), 10-meter u- and v-components of wind (U10 and V10), mean sea-level pressure (MSLP), and sea surface temperature (SST); and five atmospheric variables: geopotential (Z), temperature (T), u- and v-components of wind (U and V), and specific humidity (Q), each at $13$ pressure levels \{$50$, $100$, $150$, $200$, $250$, $300$, $400$, $500$, $600$, $700$, $850$, $925$, $1000$\} hPa. To stabilize phase shift and simplify orographic representations, we also force the model with top-of-atmosphere solar radiation, surface geopotential, and land-sea mask as input. Data from 1979--2018 are used for training, 2019 for validation, and 2020 for testing to be consistent with WB2 evaluations. The dataset totals 16~TiB in HDF5 format.

\textbf{Training } To model the stochastic dynamics of the atmosphere, we use a conditional diffusion model parameterized by TrigFlow \cite{lu2024simplifying}, which unifies EDM \cite{karras2022elucidating} and flow matching \cite{lipman2022flow,liu2022flow,albergo2023stochastic,tong2023improving} under a simpler v-prediction estimate. Given clean samples $\bm{x}_0\sim p_d$ from our data distribution, we construct noisy input samples by spherical interpolation with Gaussian noise, $\bm{x}_t = \cos(t)\bm{x}_0 + \sin(t)\bm{z}$, where $\bm{z} \sim \mathcal{N}(\bm{0}, \sigma_d^2 \bm{I})$ and $\sigma_d=1$ denotes the data standard deviation. The interpolation parameter (or diffusion time step) is defined as $t = \arctan(e^\tau/\sigma_d) \in [0, \pi/2]$ with values drawn from a prior log-uniform distribution $\tau = (1 - u)\,\log(\sigma_{\min}) + u\,\log(\sigma_{\max})$, $u \sim \mathcal{U}(0,1)$ with empirically chosen bounds of $\sigma_{\min}=0.2$ and $\sigma_{\max}=500$. This noise distribution is seen to better cover the heavy tailed distribution of target samples.

We parameterize our diffusion model as $\bm{f}_\theta(\bm{x}_t, t)=\bm{F}_\theta(\bm{x}_t / \sigma_d, t)$ where $\bm{F}_\theta$ is our network with distributed parameters $\theta$. Under TrigFlow, we estimate the target velocity $\bm{v}_t = \cos(t)\bm{z} - \sin(t)\bm{x}_0$ with the following objective
\begin{equation}\label{eq:loss.diff}
\ell^{\mathrm{Diff}}\left(\theta\right) = \mathbb{E}_{\bm{x}_0,\bm{z},t} \left[\left\Vert \sigma_d \bm{F}_\theta \left(\tfrac{\hat{\bm{x}}_t}{\sigma_d}, t\right) - \bm{v}_t \right\Vert^2_2 \right],
\end{equation}
where $\hat{\bm{x}}_t$ is a sample conditioned input. With a slight abuse of notation, let the sample at forecast time $i$ be $\bm{x}_i$. Our model estimates the target residual $\bm{x}_0 = \bm{x}_i - \bm{x}_{i-1}$ such that $\hat{\bm{x}}_t = [\bm{x}_t,\bm{x}_{i-1},\bm{x}_f]$ has initial condition $\bm{x}_{i-1}$ and forcings $\bm{x}_f$ at $i-1$ concatenated channel-wise. Data are z-score standardized with per-variable training statistics and predictions are unstandardized and added to $\bm{x}_{i-1}$ to recover the full-field state $\bm{x}_i$. 

Considering the large number of prognostic variables, we modify the above objective to a more meaningful, physically weighted loss by leveraging a latitude- and pressure-weighting, as in prior works \cite{nguyen2024scaling,lam2022graphcast}, to account for the non-uniformity of the re-gridded sphere and to emphasize near-surface variables that are most important for weather forecasting. The functions $\alpha(s)$ and $\kappa(v)$ represent the latitude and variable weights for each variable $v\in\mathcal{V}$ in our objective
\begin{equation}
\mathcal{L}(\theta) = \frac{1}{|\mathcal{S}|} \sum_{s \in \mathcal{S}} \sum_{v \in \mathcal{V}} \kappa(v) \alpha(s) \ell^{\mathrm{Diff}}_{v,s}(\theta),
\end{equation}
where $\mathcal{S}$ is the set of spatial indices over all batches. 

In our distributed setting with shared input windows loaded independently across ranks, we need to ensure noise levels are consistent per-sample. We achieve this by sharing a random seed for the state $t$ in interpolant generation for all ranks in model parallel (i.e., SP, PP, and WP), but not data parallel. The Gaussian noise $\bm{z}$ is spatially uncorrelated, sharing no seed, and is truly random across ranks.

We train all models with AdamW ($\beta=[0.85,0.9]$, $\epsilon=1e-8$, and weight decay of $\lambda=0.01$). The learning rate peaks at $5\text{e-}4$ following a linear warmup over $50$k images, then remains constant until decaying linearly to zero over the final $100$k of $3$m total images. We maintain an exponential moving average (EMA) of model parameters with a $100$k image halflife, using only these weights during inference.

\textbf{Inference } The learned dynamics are governed by the corresponding probability flow ordinary differential equation (PF-ODE) $\tfrac{\mathrm{d}\bm{x}_t}{\mathrm{d}t} = \sigma_d \bm{F}_\theta({\bm{x}_t}/ \sigma_d, t)$, which describes the evolution of a sample under the trained model. To generate a single forecast step, we numerically integrate this with 10 steps of a second-order DPMSolver++ 2S solver \cite{lu2025dpm}, modified using a log-uniform schedule for $t$ to match the training prior. Under TrigFlow, we further introduce a trigonometric Langevin-like churn that temporarily injects noise to improve sample quality and ensemble spread. New ensemble members are generated by resampling noise $\bm{z}$ at $t=\pi/2$, with each output serving as the initial condition for the next autoregressive step.

\textbf{Evaluation } We evaluate \name{} using domain preferred diagnostics under medium-range and seasonal scales. In the former we compare forecasts of a subset of variables to baseline models used in WB2. See Section \ref{ssec:domain-results} for details.

\subsection{Software Environment}

\name{} is implemented using PyTorch  
which natively supports Intel GPUs where several key kernels are optimized using Intel's extension for PyTorch (IPEX)~\cite{ipex} for the GPUs. A key enabler of distributed AI at scale on Aurora is the \textit{oneAPI Collective Communications Library} (oneCCL)~\cite{oneCCL}. Built atop $X^{e}$-Links for intra-node communication and Slingshot for inter-node connectivity, oneCCL provides a highly optimized and flexible set of collective primitives, including \texttt{broadcast}, \texttt{reduce}, \texttt{allreduce}, \texttt{allgather}, and others—tailored for Intel's XPU architecture. oneCCL is integrated with all major deep learning frameworks such as PyTorch, TensorFlow, DeepSpeed, among others. 

On LUMI we leverage the AMD software stack, which leverages ROCm backend for PyTorch, and the RCCL communication library for efficient collectives.

SWiPe leverages DeepSpeed Ulysses~\cite{deepspeedulysses} from DeepSpeed as a building block. Pipeline parallelism, data parallelism, and a Zero1-like distributed optimizer were designed using custom-built modules we developed with PyTorch. This path enabled us to tightly integrate and optimize the various parallelisms involved in \name{}, allowing us to minimize the many data transfer overheads faced with existing implementations. 

On Aurora, the SWiPe communication is overlapped with computation by offloading the communication to the CPU and then communicated with MPI-based libraries. 
This does not introduce additional latency, as the NICs on Aurora are connected to the CPU, requiring that all messages traverse it in any case.
Unfortunately, we were not yet successful in implementing this overlap on other systems as we faced hangs and poor performance with MPI-based communication, and deadlocks with *CCL based communication. The deadlocks can happen due to the asynchronous p2p communication that is executed on a stream. This implies a modified pipeline schedule with coupled send and receive calls is required to take advantage of the stream-based deep learning-focused communication libraries. 

We primarily used the \texttt{BF16} floating point format for most of our computational kernels. The embeddings, gradients, parameters, and gradient reduction were done in single precision floating point format (\texttt{FP32}) for numerical stability. As a note, \texttt{FP32} computations were just a very small fraction of the overall compute as most compute work is in the forward and backward pass with \texttt{BF16} activations and model weights. We used a PyTorch dataloader using h5py to load the datasets. Our implementation of \name{} is portable and can run on any system or supercomputer that supports PyTorch as demonstrated by our evaluations on two different accelerators from Intel and AMD. 

\subsection{Measurement Methodology}

We develop an analytical model to estimate floating point operations, which takes into account various \name{} model parameters. This model builds on our prior work in accurately modeling transformers in Megatron-DeepSpeed at large scale \cite{dharuman2024mprot}. The sustained FLOPS were 
determined by measuring the end-to-end time for the entire training loop. This included the I/O time needed to load and pre-process the data, communication time, including time spent in model-parallelism, gradient synchronization, among others. The peak FLOPS were determined 
by measuring only the time taken for the compute-intensive part of the training loop - executing the full pipeline schedule and accounts for all the communication time needed for SWiPe; it does not account for the time spent on gradient synchronization. 

\begin{figure*}[!t]
    \centering
    \begin{subfigure}[b]{0.49\textwidth}
        \centering
        \includegraphics[width=\linewidth]{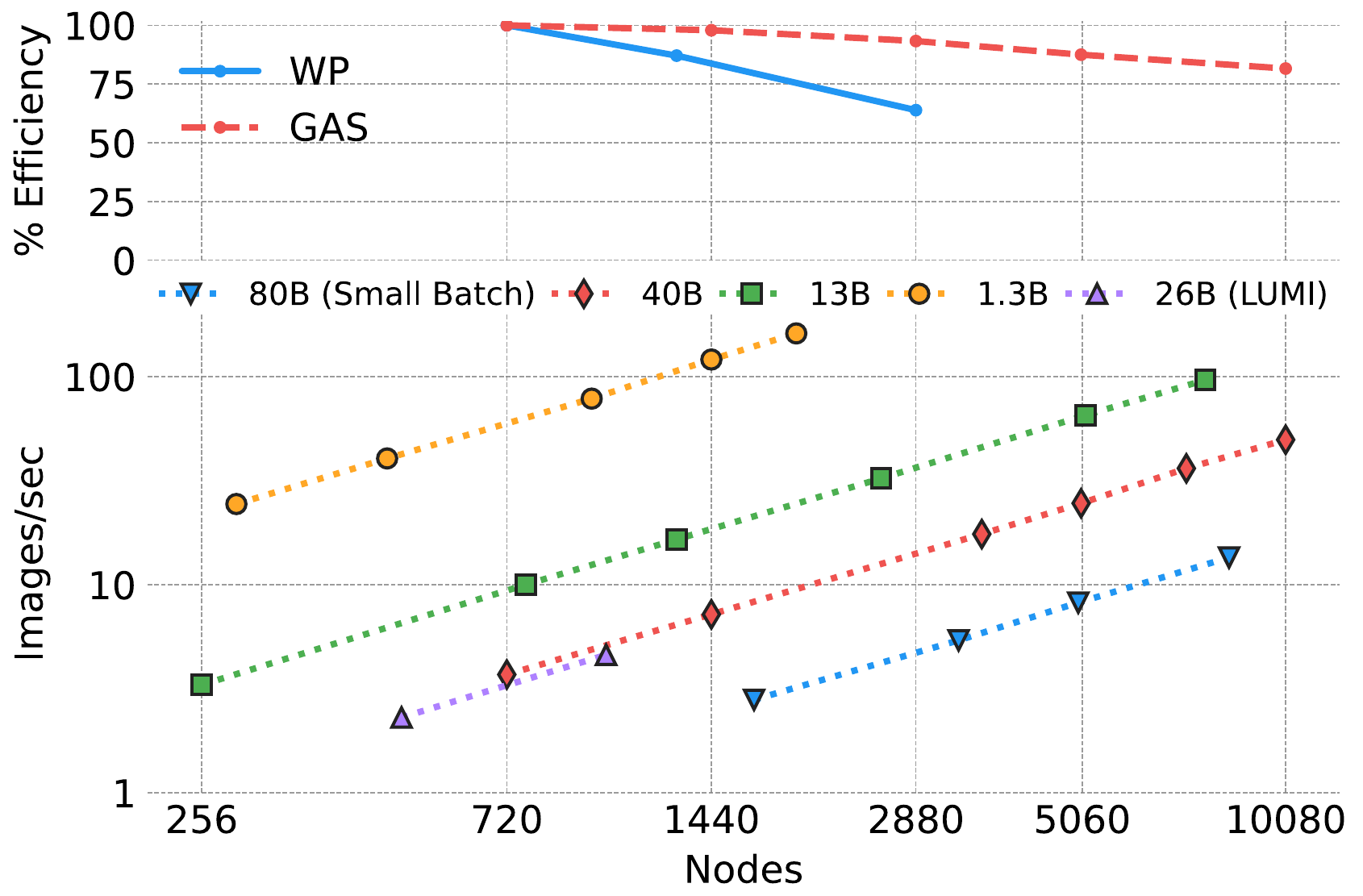}  
        \caption{Strong scaling efficiency (top), images/sec weak scaling (bottom).}
        \label{fig:scaling_plot.strong}
    \end{subfigure}
    \hfill
    \begin{subfigure}[b]{0.49\textwidth}
      \centering
      \includegraphics[width=\linewidth]{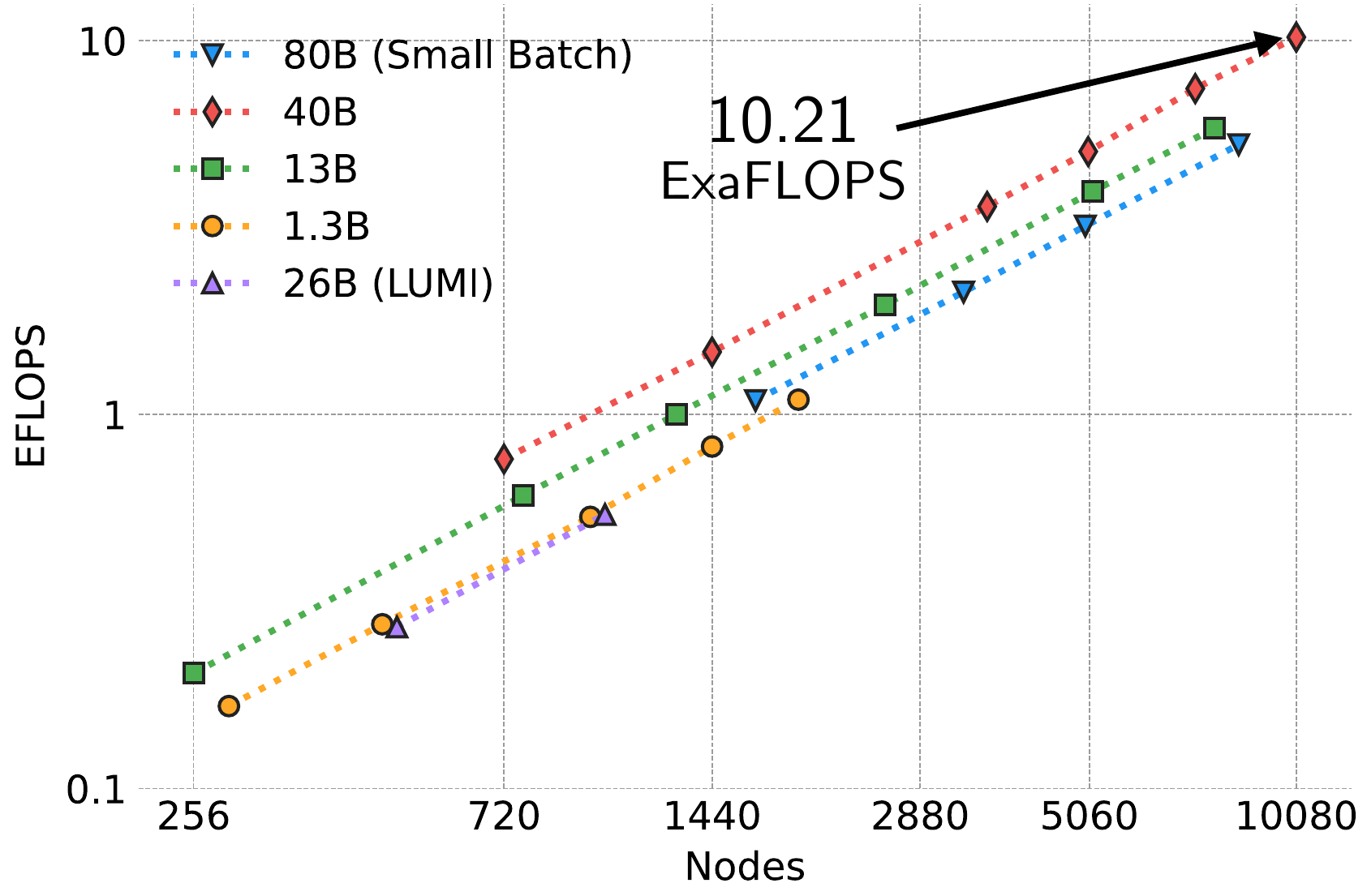}  
      \caption{Weak scaling performance---sustained FLOPS.}
      \label{fig:scaling_plot.weak}
    \end{subfigure}
  \caption{Strong and weak scaling of training on Aurora (and LUMI) for multiple configurations with fixed ($\text{WP} \times \text{PP} \times \text{SP}$) in log-log scale. Strong scaling in the top for the 40B configuration is driven by changing either gradient accumulation steps (GAS) or window parallelism degree (WP); and the weak scaling is driven by increasing data parallelism, demonstrating consistent exascale performance and high utilization across large node counts.}
  \label{fig:scaling_plot}
\end{figure*}

\section{Performance Results}\label{sec:perf} 

We assess the scalability of the \name{} model training on Aurora, followed by a comparative evaluation of its performance across a variety of forecasting tasks against state-of-the-art models.

\subsection{Computational Performance}

We evaluate the scaling performance of four \name{} model configurations as depicted in Table ~\ref{tab:config-table}. This represents models with increasing number of parameters ranging from 1.3 Billion to 80 Billion. The table includes the window parallelism (WP), pipeline parallelism (PP), and sequence parallelism (SP) used, as well as the model parameters - hidden dimensions, number of heads and feed-forward network size - for each. As mentioned previously in Section~\ref{sec:innovation}, we target sequence parallelism within a node to use all 12 GPU tiles on an Aurora node and 8 GPU tiles on LUMI node to optimize the performance by restricting the communication within the node. The number of nodes needed to run a single model instance is given by $WP \times PP$. The LUMI 26B configuration differs from the Aurora 40B configurations mainly due to the reduced sequence parallelism, which is balanced by increased window parallelism. We also had to reduce the number of pipeline stages to fit the configuration to the standard queue of 1024 nodes. We weak scale these configurations using data parallelism to the node counts, parallelism and global batch size depicted in Table \ref{tab:perf_results}.

\begin{table}[!t]
 \footnotesize
 \caption{\name{} model configurations.}
 \centering
 \begin{tabular}{r|ccccccc}
 \toprule
 Params      & WP  & PP & GAS &  Dim & Heads & FFN  & Nodes \\
 \midrule
 1.3B       & 4($2\times2$) & 12 & 60 & 1536      & 12    & 9216 & 48 \\
 13B         & 16($4\times4$) & 16 & 48 & 4608      & 36    & 25600 & 256\\
 40B        & 16($4\times4$) & 20 & 140 & 6144      & 48    & 40960 & 720 \\
 80B        & 36($6\times6$) & 26 & 52 & 7680      & 60    & 46080 & 1664\\
 \midrule
26B(L)        & 36($6\times6$) & 14 & 70 & 6144      & 48    & 32768 & 504\\
 \bottomrule
 \end{tabular}
 \label{tab:config-table}
\vspace{-5mm}
\end{table}

Figure \ref{fig:scaling_plot} depicts the weak scaling performance achieved in terms of throughput in images/sec as we increase the data parallelism to scale on Aurora. From the figure, we observe nearly linear scaling as we weak scale for all the model configurations on Aurora. 
In general, at similar node count, we achieve higher throughput with larger models in comparison to smaller models due to the increase computational requirements. The exception is the 79B configuration, as that one has significantly smaller batch size. 

At the scale of 1440 nodes, we observe a 18$\times$ improvement in throughput for the 1.3B model over the 40B model, which is $31.5\times$ larger. This can be attributed to the underlying Model FLOPS Utilization (MFU) in Table \ref{tab:perf_results} due to the lower compute to communication ratio with respect to the 40B model at this scale.


Figure \ref{fig:scaling_plot} also demonstrates 
the large-scale performance of training \name{} on Aurora achieving \textbf{sustained multi-exaflop training} in mixed-precision at scale. Notably, for the 40B \name{} configuration with a WP=36 and PP=20, we attain a \textbf{sustained performance of 10.21 ExaFLOPS}, and a \textbf{peak performance of 11.21  ExaFLOPS}, marking the highest throughput observed across all configurations. These results are enabled by a large degree of parallelism, combining domain-decomposition via window parallelism and sequence parallelism, and pipeline model parallelism, yielding an overall parallel degree of $36\times 20\times12$. This parallel strategy allows \name{} training at a large scale without a large data parallelism degree, thus, leading to more stable training. At full scale of 10,080 nodes, the 40B model achieved a throughput of 50 samples per second. At this pace, it would take approximately 15 hours to complete training for 3M samples. 

We also demonstrate an extreme case of scaling the model size and parallelism by presenting the 80B parameter configuration with WP=64 and smaller gradient accumulation, achieving \textbf{5.27 ExaFLOPS with a global batch size of just 260 samples at 8320 nodes}, resulting in unprecedented model and input parallelism requiring just one sample per 384 GPU tiles. In practice, the model parameters and the individual samples are sharded further, but more samples pass through a single GPU. 

\begin{figure*}[!t]
  \centering
  \includegraphics[width=\textwidth]{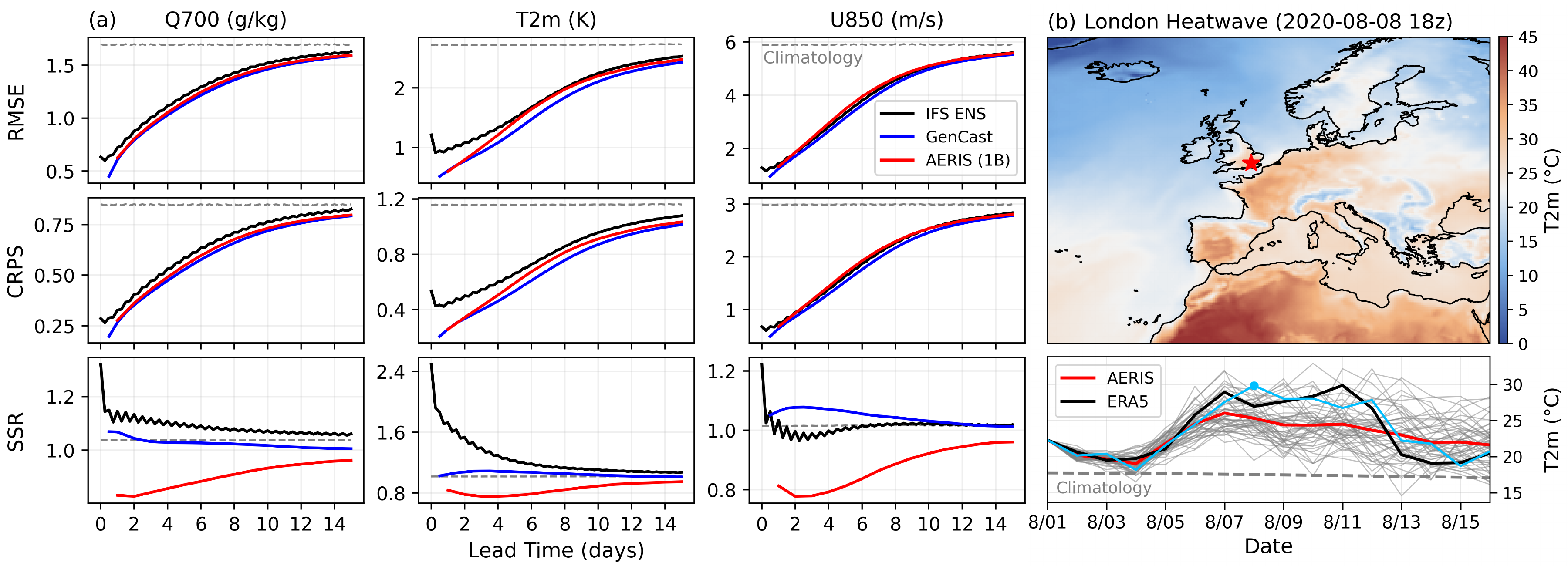}
  \caption{Medium-range forecast skill. (a) Latitude-weighted root-mean-squared error (RMSE), continuous ranked probability score (CRPS), and spread/skill ratio (SSR) across 3 key variables for 50 ensembles of 155 initial conditions; and (b) accurate heat wave forecast over London, England (\textcolor{red}{red star}), initialized 2020-08-01T18z, showing 50 \textcolor{gray}{ensembles} and the \textcolor{cyan}{closest} member.}
  \label{fig:rmse_results}
\end{figure*}

Our approach enables \name{} to scale efficiently across the full Aurora system. The compute throughput scaling shown in the bottom side of Figure \ref{fig:scaling_plot} highlights the weak scaling efficiency as we increase the number of nodes along the data parallel dimension, and thus the batch size, under fixed model-parallel settings. In particular, the 40B (WP=36, PP=20) configuration demonstrates good weak scaling efficiency of 95\%, maintaining high throughput across 10,080 nodes of the Aurora production system. On both systems, we achieve a mean 
Model FLOPS Utilization (MFU) 
of over 30\%, indicating a good overall utilization. The utilization is observed to be limited by typical reasons such as computational kernel efficiency, hard-to-hide communication overhead, and the idle time caused by the pipeline bubble. The performance achieved showcases the ability of our hybrid parallel architecture to support training at extreme scales, making \name{} a compelling foundation for future scientific AI workloads. 

The top side of Figure \ref{fig:scaling_plot} shows the strong scaling properties when adjusting the gradient accumulation steps (GAS) or window parallelism (WP) degree to get the same batch size (1960 for GAS scaling, 140 for WP scaling), and thus an equivalent training step across the range. The GAS scaling achieves strong scaling of 81.6\%. The losses are mainly from the increasing pipeline bubble. The WP scaling goes through WP of 36, 64, and 144, with scaling efficiencies 100\%, 87\%, and 64\%. The loss of efficiency is because the reduced GPU saturation due to less data per GPU, and relatively larger portion of time spent on gradient reduction, overall reducing GPU utilization. WP=144 is $4\times$ larger than WP=36, but only achieves $2.4\times$ speedup, resulting in strong scaling of 64\% in the extreme case of 140 samples on 2880 nodes without data parallelism.

For achieving the performance, we also had to optimize our data loading and processing pipelines. To realize this, we isolate the input/output embedding layers and data I/O into separate stages resulting in a reduction of the pipeline bubble. Combining I/O and embeddings within the main pipeline stages would introduce additional latency that propagates as pipeline bubbles across all stages. By separating them out, we localize the impact to the first and last stages, resulting in slightly reduced GPU utilization at the edges but significantly better efficiency across the full pipeline. With this design, the number of pipeline stages is $PP = L + 2$ where $L$ is the number of Swin layers. 
\begin{figure*}[!t]
  \centering
  \includegraphics[width=\linewidth]{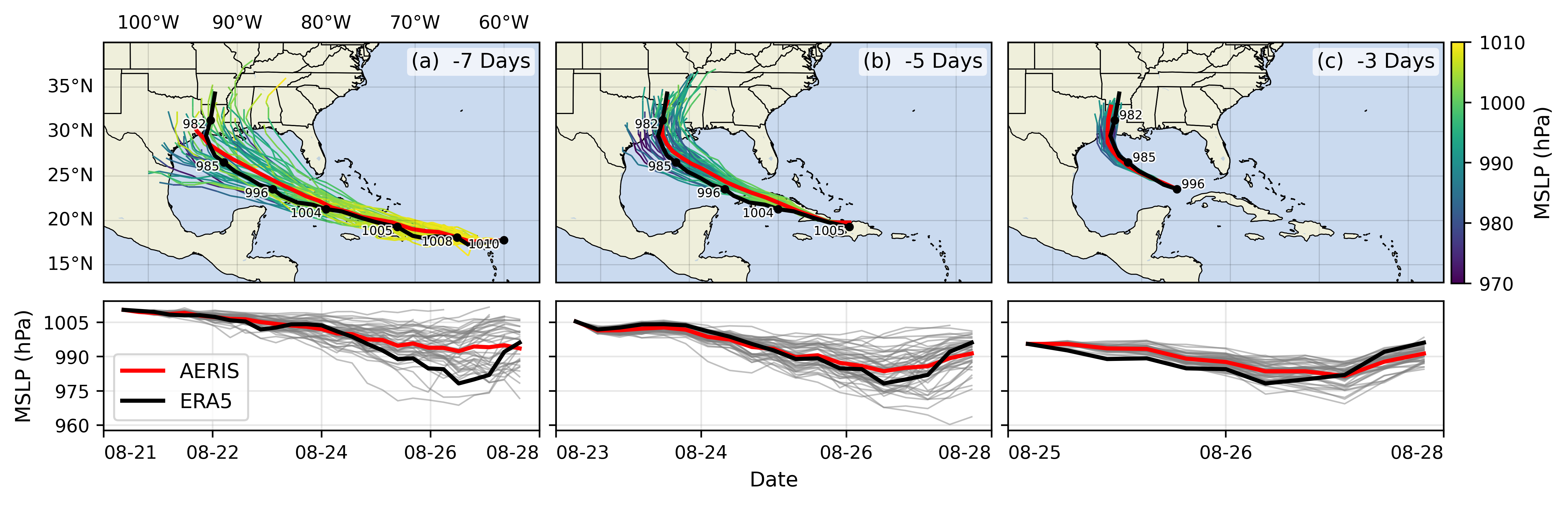}
  \caption{Hurricane Laura tracks (top) and intensity (bottom). Initialized 7(a), 5(b), and 3(c) days prior to 2020-08-28T00z.}
  \label{fig:hurricane}
  \vspace{-5mm}
\end{figure*}

\begin{table}[!t]
 \small \caption{\label{tab:model_arch} Sustained and peak training throughput for \name{} on Aurora (and LUMI for 26B) across different model sizes: DP -- Data Parallel degree, GBS -- Global Batch Size, 
 MFU -- 
 Model FLOPS Utilization, 
 TF/T -- TFLOPS per tile, EF(S) -- sustained ExaFLOPS, and EF(P) -- peak ExaFLOPS. The gap between peak and sustained ExaFLOPS is primarily due to the time spent on the optimizer step and gradient reduction. These components occur outside the pipelined forward-backward pass and thus contribute to the reduction in sustained throughput relative to the peak.
 }
 \footnotesize
 \centering
 \begin{tabular}{r|ccccccc}
 \toprule
Config & Nodes & DP    & GBS   & TF/T & MFU(\%) & EF(S)  & EF(P) \\  
\midrule
 1.3B   & 1920  & 40    & 2400  & 47.6   & 21.6 &   1.1        & 1.2      \\
 13B    & 7680  & 30    & 1440  & 63.3   & 28.8 &   5.8        & 6.4      \\
 40B    & \textbf{10080}  & 14    & 1960  & \textbf{84.4}&\textbf{38.4}&\textbf{10.21}& \textbf{11.21} \\
 80B       & 8320 & 5     & \textbf{260} & 52.8   &24 &   5.27        & 6.1      \\
 \midrule
 26B(L)    & 1008 & 2   & 140 & 66.5   &34.8 &   0.54        & 0.62      \\
 \bottomrule
 \end{tabular}
 \label{tab:perf_results}
 \vspace{-5mm}
\end{table}

\subsection{\label{ssec:domain-results}Domain Results}
A core innovation of \name{} is in being an end-to-end generative model for short-term weather forecasting to seasonal scales. We demonstrate this by evaluating skill on medium-range weather forecasting (lead-times of 1--14 days), performance on notable features that are predictable on the subseasonal-to-seasonal (S2S) time scales, e.g., sea surface temperature (SST) and Madden--Julian oscillation (MJO), and lastly explore performance on extreme events including record breaking heatwaves and tropical cyclones. 

\textbf{Medium-range forecasting } 
For medium-range weather forecasting, we assess the ability of our model to produce well calibrated and skillful probabilistic forecasts. We compare to two well-established models: (1) GenCast \cite{price2024gencast}, a diffusion-based ensemble system from Google DeepMind; and (2) the IFS ensemble (IFS ENS) a state-of-the-art, numerical-based ensemble system from the European Center for Medium Range Weather Forecasting (ECMWF) \cite{lang2023ifs}. Evaluations are conducted on held-out, test data for 2020, where we showcase ensemble mean latitude-weighted root-mean-squared error (RMSE), a probabilistic metric called Continuous Ranked Probability Score (CRPS), and the spread/skill ratio (SSR) for a subset of variables over 14-day rollouts (on 24h intervals). 

Figure \ref{fig:rmse_results}a illustrates our results for each of these variables. We find \name{} is on-par or outperforms the IFS ENS for ensemble mean RMSE and CPRS, while performing competitively with GenCast especially during the 1--3 day and 10$+$ day time frames. In fact, for specific humidity at 700 hPa, \name{} is nearly identical to GenCast in terms of forecast performance. This is despite \name{} being an under-dispersive ensemble system (SSR $< 1$), suggesting the potential for improvement in model performance by increasing the diversity and spread of individual ensemble members (see Section~\ref{ssec:futurework}). This result of using only diffusion for ensemble members leading to under-dispersion is not unique as this was also found to be true in GenCast.

\begin{figure*}[!th]
  \centering
  \includegraphics[width=\linewidth]{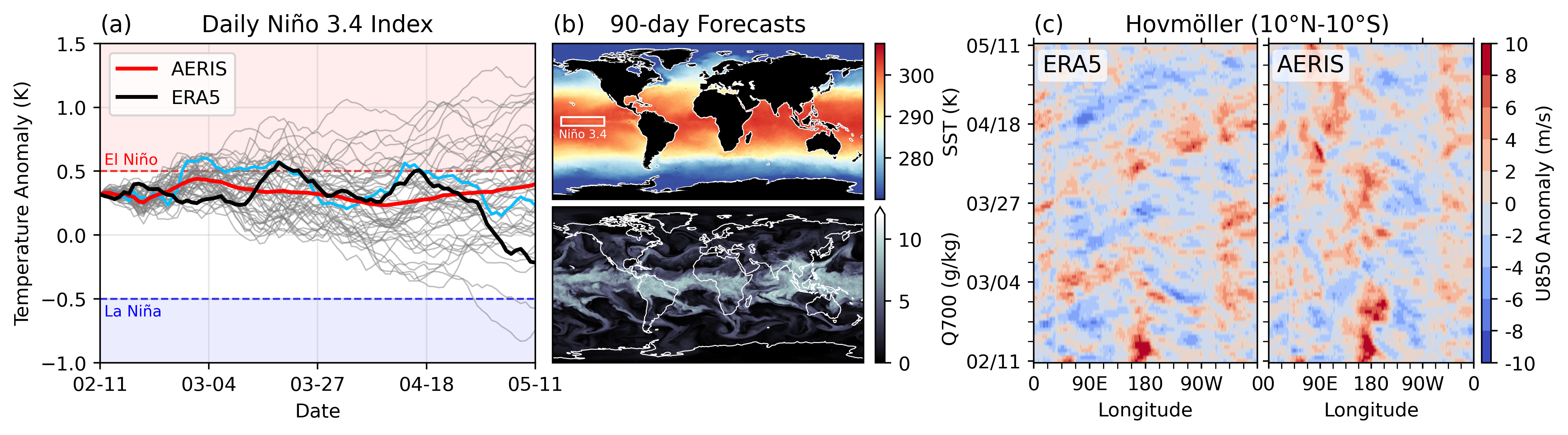}
  \caption{Seasonal forecast stability. (a) Spring barrier under El Ni\~no with realistic ensemble spread in the ocean; (b) qualitatively sharp fields of SST and Q700 predicted 90 days in the future from the \textcolor{cyan}{closest} ensemble member to ERA5 in (a); and (c) stable Hovm{\"o}ller diagrams of U850 anomalies (climatology removed; m/s), averaged between $10^\circ$N and $10^\circ$S for a 90-day rollout.}
  \label{fig:s2s}
  \vspace{-2mm}
\end{figure*}

\textbf{Subseasonal-to-seasonal (S2S) } 
To assess trends that extend beyond that of our medium-range weather forecasts (beyond 14-days) and evaluate the stability of our model, we made $3{,}000$ forecasts (60 initial conditions each with 50 ensembles) out to 90 days. \name{} was found to be stable during these 90-day forecasts with realistic atmospheric states (Figure \ref{fig:s2s}b), and correct power-spectra even at the smallest scales (not shown). We demonstrate for the first time, the ability of a generative, high-resolution (native ERA5) diffusion model to produce skillful forecasts on the S2S timescales with realistic evolutions of the Earth system (atmosphere + ocean).

We showcase this by evaluating the model's ability to predict the El Ni\~no Southern Oscillation (ENSO) state by demonstrating skillful predictions of daily Ni\~no 3.4 indices out to at least 90 days during the winter and spring months in 2020 (Figure \ref{fig:s2s}a). We find the ensemble mean mirrors the true evolution of the equatorial Pacific SSTs for 90 days with realistic spread along the spring barrier. We also find on a global scale, realistic error growth and forecast skill on-par with numerical-based coupled systems (not shown). Lastly, we look at a brief study of how convectively coupled equatorial waves propagate through longitude and time. More specifically, in Figure \ref{fig:s2s}c, we qualitatively compare Hovm{\"o}ller diagrams \cite{hovmoller1949trough} of a single ensemble, seeing skill to at least 3 weeks and shows realistic variability out to 90 days. We emphasize that to our knowledge, this is the first diffusion-based model for atmosphere and ocean prediction that demonstrates skill on the S2S timescales at $0.25^\circ$.

\textbf{Extreme Events} 
The correct predictions of extreme events (e.g., tropical cyclones and heatwaves), are vital for weather forecasting due to their high socioeconomic impacts and their potential to cause large loss of life. With its probabilistic formulation, we find that \name{} is able to predict extreme events with exceptional skill. To demonstrate this, we examine two case studies: (1) correctly predicting Hurricane Laura's track and intensity up to 7 days before landfall; and (2) the prediction of the record-breaking heatwave in Europe during early August of 2020 (Figure ~\ref{fig:rmse_results}c and ~\ref{fig:hurricane}).

Hurricane Laura made landfall near Cameron, Louisiana, on August 27, 2020, causing \$19 billion in damage and 47 direct deaths \cite{pasch2021hurricane}. This record-breaking hurricane is well forecasted by \name{} with minimal track errors even with a lead-time of 7 days. By day 5, the ensemble mean correctly captures the eventual northward track and has landfall in Louisiana, nearly identical to the ERA5 track. Another aspect of Hurricane Laura is the rapid intensification in the Gulf of Mexico. Our model accurately predicts this intensification period  with a lead time of 5 days, a phenomena that is not typically well captured by global numerical models.

During August 2020, an intense heat wave impacted Europe, with the United Kingdom being the most affected. We show\name{} successfully identifying this heatwave with a lead time of more than a week (Figure ~\ref{fig:rmse_results}b). Specifically, all ensemble members capture the sharp rise in temperatures, followed by the return to climatology, with the ensemble mean closely following ERA5. These results signify the ability of \name{} in positioning itself as a reliable and skillful model, even at the tail of distributions, where extreme events occur.

\subsection{\label{ssec:futurework}Limitations and Future Works}

Despite strong results, \name{} faces several limitations that we aim to address. Our science findings are based on our 1.3B parameter model, and while we are currently pursuing highly ambitious trainings (e.g., 13B with $1\times1$ patch-size), these experiments require extensive compute resources to converge---on the order of one week of run-time on exascale machines ($\sim$1.5M node hours)---which poses practical challenges. Beyond compute constraints, current medium-range skill (Figure~\ref{fig:rmse_results}) remains overconfident. Improving the spread/skill ratio through initial condition perturbations and tuning our stochastic churning schedule under TrigFlow may improve ensemble spread without hurting skill. Our diffusion parameterization also allows for consistency distillation \cite{lu2024simplifying}, which allows us to compress the model size and reduce inference to a single step, thereby lowering computational cost by orders of magnitude for generating new forecasts. As a consistency model, \name{} could benefit from multi-step finetuning \cite{stock2025swift}, which may yield measurable improvements to forecast skill \cite{siddiqui2024exploring}.

\name{} is data-driven and trained on historical reanalysis, not governed by numerical equations; thus, non-physical artifacts or unrealistic dynamics can arise. Additionally, to make our model operational under current conditions, finetuning on IFS HRES 0th frame data would be required. At the same time, the methods presented herein are generally adaptable and can scale to higher-resolution inputs and generalize to alternative datasets. Ongoing work explores such datasets and architectural changes to extend forecast skill to even longer time horizons. Finally, SWiPe itself can be improved by reducing the bubble size of pipeline parallelism, as GPUs currently idle when waiting for data from another pipeline stage under 1F1B; adopting zero-bubble pipeline parallelism \cite{qi2024zero} offers a promising solution.

\section{Implications}
Weather forecasting and climate modeling have been a grand challenge in science and computing for the past 50 years. Advances in computing have helped extend weather forecast skill from 3 to nearly 8 days and improved climate model resolution from 500 km in the 1980s to about 25 km today \cite{bauer2015quiet}. While this progress is impressive, this has come at a cost of expensive model simulations. Even with the latest computing capabilities, it remains a challenge to produce both large ensembles and achieve high-spatial resolution for both weather and seasonal scales. 

Today, the state-of-the-art approaches for modeling weather are inspired by the tremendous progress in artificial intelligence (AI) approaches and building observational data-based models. These models have shown tremendous progress over the past few years, and with forecast capability on short- and medium-range weather forecasting either matching or beating those produced by conventional numerical weather forecast models operated by national and international facilities. 

We demonstrate a significant advancement in AI weather and climate modeling with \name{} by efficient scaling of window-based transformer models. We have performed global medium-range forecasts with performance competitive with GenCast and surpassing the IFS ENS model, with longer, 90-day rollouts showing our ability to learn atmospheric dynamics on seasonal scales without collapsing, becoming the first diffusion-based model that can work across forecast scales from 6 hours all the way to 3 months with remarkably accurate out of distribution predictions of extreme events. 

On the system architecture side, \name{} exploits an optimized hybrid parallelism strategy---integrating our window-based transformer parallelism with pipeline, sequence, and data parallelisms---to achieve exceptional computational throughput and scaling performance. Our proposed parallelisms in SWiPe accelerates  training of our \name{} models  and enables scaling to large node counts at lower global batch size. We are able to scale to 10,080 nodes (120,960 GPU tiles) and achieve a maximum sustained performance of 10.21 ExaFLOPS in mixed precision, setting a high bar for model training performance. These innovations significantly advance the feasibility of training foundation climate models on the highest resolution data on exascale systems.
\section*{Acknowledgments}
\small{This research used resources of the Argonne Leadership Computing Facility, a U.S. Department of Energy (DOE) Office of Science user facility at Argonne National Laboratory and is based on research supported by the U.S. DOE Office of Science-Advanced Scientific Computing Research Program, under Contract No. DE-AC02-06CH11357. We acknowledge CSC – IT Center for Science, Finland, for computational resources on LUMI and thank Pekka Manninen at CSC for timely support. We thank Servesh Muralidharan from Argonne Leadership Computing Facility (ALCF) for help reducing job startup time at launch and help with debugging computational performance, Nithin Chalapathi from UC Berkeley for running experimental GenCast experiments for baselines, and Joseph Insley from Argonne National Laboratory for supporting preliminary experimental visualization efforts.}

\bibliographystyle{IEEEtran}
\bibliography{references}

\end{document}